\documentclass[final]{cvpr}

\usepackage{times}
\usepackage{epsfig}
\usepackage{graphicx}
\usepackage{amsmath}
\usepackage{amssymb}

\usepackage{url}
\usepackage{graphics}
\usepackage{graphicx}
\usepackage{layouts}
\usepackage{ulem}
\usepackage{multirow}
\usepackage{tabu,stackengine}
\usepackage{caption}
\usepackage{wrapfig}
\usepackage{floatrow}
\usepackage{booktabs}
\usepackage{soul}
\newfloatcommand{capbtabbox}{table}[][\FBwidth]
\usepackage{caption}
\usepackage{subcaption}
\usepackage{makecell}

\usepackage[pagebackref=true,breaklinks=true,colorlinks,bookmarks=false]{hyperref}

\def\cvprPaperID{34} %
\def\confYear{CVPR 2021}

\newcommand{\dc}[1]{{\color{black}#1}}

\newcommand{\op}[1]{{\color{black}#1}}

\begin{document}

\title{BalaGAN: Cross-Modal Image Translation Between Imbalanced Domains}

\author{Or Patashnik\\
Tel-Aviv University
\and
Dov Danon\\
Tel-Aviv University
\and
Hao Zhang\\
Simon Fraser University

\and
Daniel Cohen-Or\\
Tel-Aviv University
}

\twocolumn[{%
\renewcommand\twocolumn[1][]{#1}%
\maketitle
\begin{center}
    \centering
    \def\arraystretch{0.7}
    \setlength{\tabcolsep}{1.3pt}
    		\begin{tabu}{c@{\hspace{0\tabcolsep}} c c @{\hspace{0\tabcolsep}}c@{\hspace{0\tabcolsep}} c@{\hspace{0\tabcolsep}} c@{\hspace{0\tabcolsep}} c@{\hspace{0\tabcolsep}}}
            \includegraphics[height=0.123\textwidth]{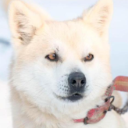} &
            \includegraphics[height=0.123\textwidth]{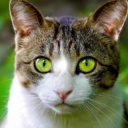} &
            \includegraphics[height=0.123\textwidth]{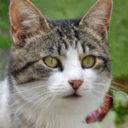} &
            \includegraphics[height=0.123\textwidth]{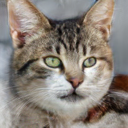} &
            \includegraphics[height=0.123\textwidth]{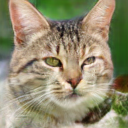} &
            \includegraphics[height=0.123\textwidth]{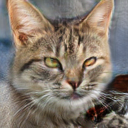} & \includegraphics[height=0.123\textwidth]{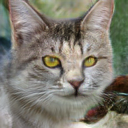} \\

            \includegraphics[height=0.123\textwidth]{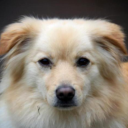} &
            \includegraphics[height=0.123\textwidth]{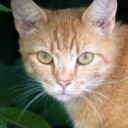} &
            \includegraphics[height=0.123\textwidth]{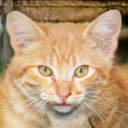} &
            \includegraphics[height=0.123\textwidth]{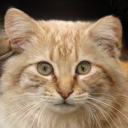} &
            \includegraphics[height=0.123\textwidth]{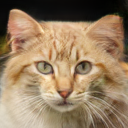} &
            \includegraphics[height=0.123\textwidth]{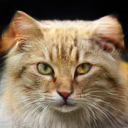} & \includegraphics[height=0.123\textwidth]{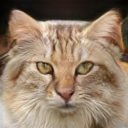} \\
            \small{Source} & \small{Reference} & \small{Full Domain} & \small{1000} & \small{500} & \small{250} & \small{125}
            
        \end{tabu}
    \vspace{-7pt}
    
    \captionof{figure}{
    Our network, BalaGAN, translates a dog to a cat, based on a reference image. We train the network on 4739 dog images and decreasing number of cat images, from full domain (5153 cats) down to 125, leading to more and more imbalanced domain pairs. Quality of the translated images remains high, even when the two domains are highly imbalanced.
    }
\end{center}

}]

\begin{abstract}
    State-of-the-art image translation methods tend to struggle in an imbalanced domain setting, where one image domain lacks richness and diversity.
We introduce a new unsupervised translation network, BalaGAN, specifically designed to tackle the domain imbalance problem. We leverage the latent modalities of the richer domain to turn the image-to-image translation problem, between two imbalanced domains, into a multi-class translation problem, more resembling the style transfer setting.
Specifically, we analyze the source domain and learn a decomposition of it into a set of latent modes or classes, without any supervision. This leaves us with a multitude of balanced cross-domain translation tasks, between all pairs of classes, including the target domain.
During inference, the trained network takes as input a source image, as well as a reference style image from one of the modes as a condition, and produces an image which resembles the source on the pixel-wise level, but shares the same mode as the reference.
We show that employing modalities within the dataset improves the quality of the translated images, and that BalaGAN outperforms strong baselines of both unconditioned and style-transfer-based image-to-image translation methods, in terms of image quality and diversity.
    
\end{abstract}

\section{Introduction}
\label{sec:intro}

\begin{figure*}[h!]
    \begin{center}
    \includegraphics[width=\textwidth]{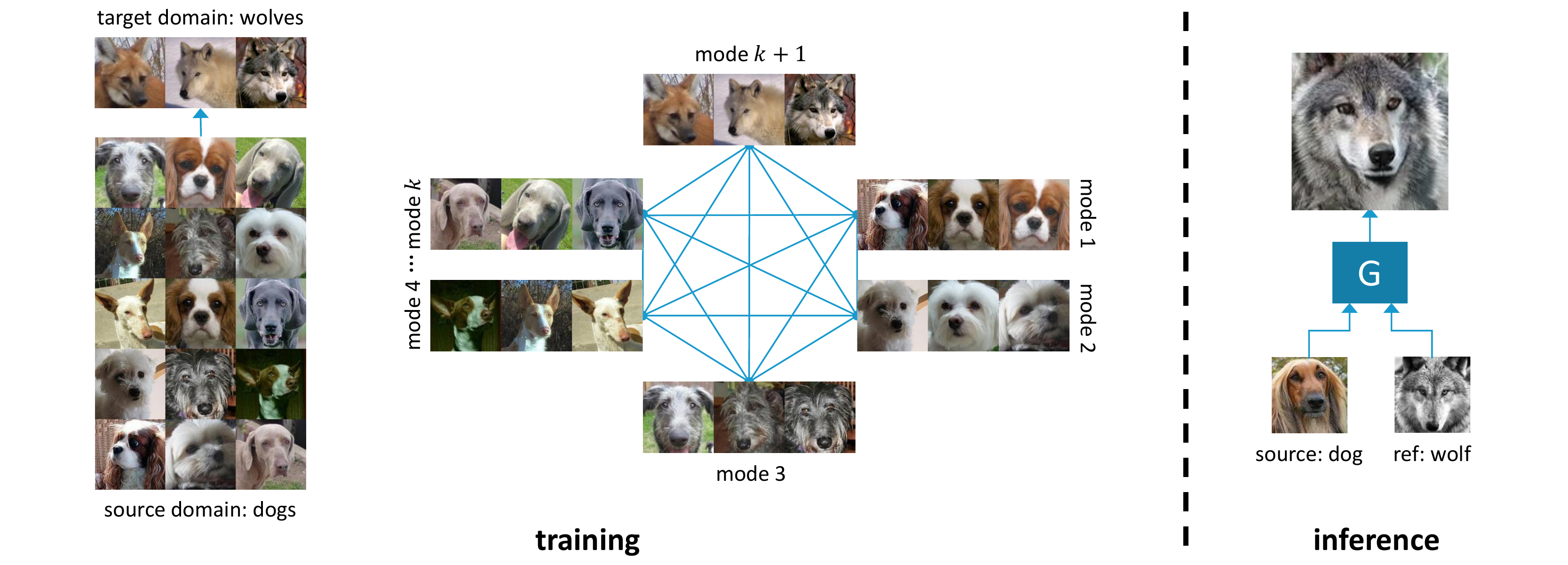}
    \end{center}
    \caption{Our image translation network, BalaGAN, is designed to handle imbalanced input domains, e.g., a set of dog images that is much richer than that of wolves. 
    \op{We decompose the source domain}
    into multiple modes reflecting the images ``styles'' and train a GAN over \textit{all mode pairs} to learn a multitude of intra- and inter-mode cross-translations. 
    During inference, the network takes a source (e.g., a dog) and a reference image (e.g., a wolf) to produce a new image following the ``style/mode" of the reference while resembling the source in a pixel-wise manner.}
    \label{fig:overview}
\end{figure*}

Image-to-image translation is a central problem in computer vision and has a wide variety of applications including image editing, style transfer, data enrichment, image colorization, etc. 
Acquiring labeled pairs of source and target domain images is often hard or impossible, thus motivating the development of unsupervised methods \cite{CycleGAN2017, huang2018munit, Kim2020UGATITUG, park2020cut, GANHopper2020, liu2019few, choi2020starganv2}.
However, these methods are often lacking in quality or robustness to domain variations.
Indeed, in most unsupervised approaches, there is an implicit assumption of ``approximate symmetry'' between the translated domains, in term of data quantity or variety.
With this assumption, the source and target domains are treated each as \textit{one-piece}, without fully leveraging the variety within either of them.
In reality, most datasets are imbalanced across different categories, e.g., ImageNet \cite{imagenet_cvpr09} contains many more images of dogs than of wolves. %
As image-to-image translation can be used to enrich some domains by utilizing others, improving these methods, in the imbalanced setting in particular, can play a critical role in resolving the ubiquitous ``data shortage'' problem in deep learning.

In this paper, we present \textit{BalaGAN}, an \textit{unsupervised} image-to-image translation network specifically designed to tackle the domain imbalance problem where the source domain is much richer, in quantity and variety, than the target one.
Since the richer domain is, in many cases, \textit{multi-modal}, we can leverage its \textit{latent} modalities. To do this, we turn the image-to-image translation problem, between two imbalanced domains, into a \textit{multi-class} and \op{\textit{reference-guided}} translation problem, akin to style transfer.
Our key observation is that the performance of a domain translation network can be significantly boosted by (i) disentangling the complexity of the data, as reflected by the natural modalities in the data, and (ii) training it to carry out a multitude of varied translation tasks instead of a single one. BalaGAN fulfills both criteria by learning translations between \textit{all pairs} of source domain modalities \textit{and} the target domain, rather than only between the full source and target domains. This way, we are taking a \op{more balanced} view of the two otherwise imbalanced domains. More importantly, enforcing the network to learn such a richer set of translations leads to improved results, and in particular, a better and more diverse translation to the target domain.

Specifically, let us assume that the source domain $A$, which is significantly richer than the target domain $B$, consists of multiple mode classes. We train a \textit{single} GAN translator $G$ with respect to all pairs of modes (see Figure~\ref{fig:overview}). During inference, the trained network takes as input a source image $x$, as well as a \textit{reference image} $y$ from one of the modes as a condition, and produces an image $G(x,y)$. This image resembles $x$ on the \textit{pixel-wise} level, but shares the same mode (or style) as $y$.
To realize our approach, we develop means to find the latent data modalities without any supervision and a powerful generator for the task of conditional, multi-class image-to-image translation. 
Our translator is trained adversarially with two discriminators, each aiming to classify a given image to its corresponding mode, with one trained on real images only. The generator is trained to produce meaningful content and style representations, and combine them through an AdaIN layer.
While this architecture bears resemblance to multi-class translation networks such as FUNIT~\cite{liu2019few} and StarGAN~\cite{choi2020starganv2}, it should be emphasized
that unlike these methods, we learn the latent modalities, and use \textit{transductive learning}, where the target domain
participates in the training.

We show that reducing the imbalanced image translation problem into a cross-modal one achieves comparable or better results compared to any unsupervised translation method we have tested, including the best performing and most established ones, since they do not exploit the latent modalities within the source domain. 
We analyze the impact of the extracted latent modalities, perform ablation studies, and extensive quantitative and qualitative evaluations, which are further validated through a perceptual user study. We further show the potential of our cross-modal approach for boosting the performance of translation in balanced setting.  

\op{Our main contributions can be summarized as follows: (i) we present a unique solution to the image translation problem that is tailored to the imbalanced domain setting; (ii) we introduce a new approach which converts a single translation problem into a multitude of cross-modal translation problems; and (iii) we demonstrate competence of our approach also under the general domain setting.}

\begin{figure*}[h]
\begin{center}
\includegraphics[width=\textwidth]{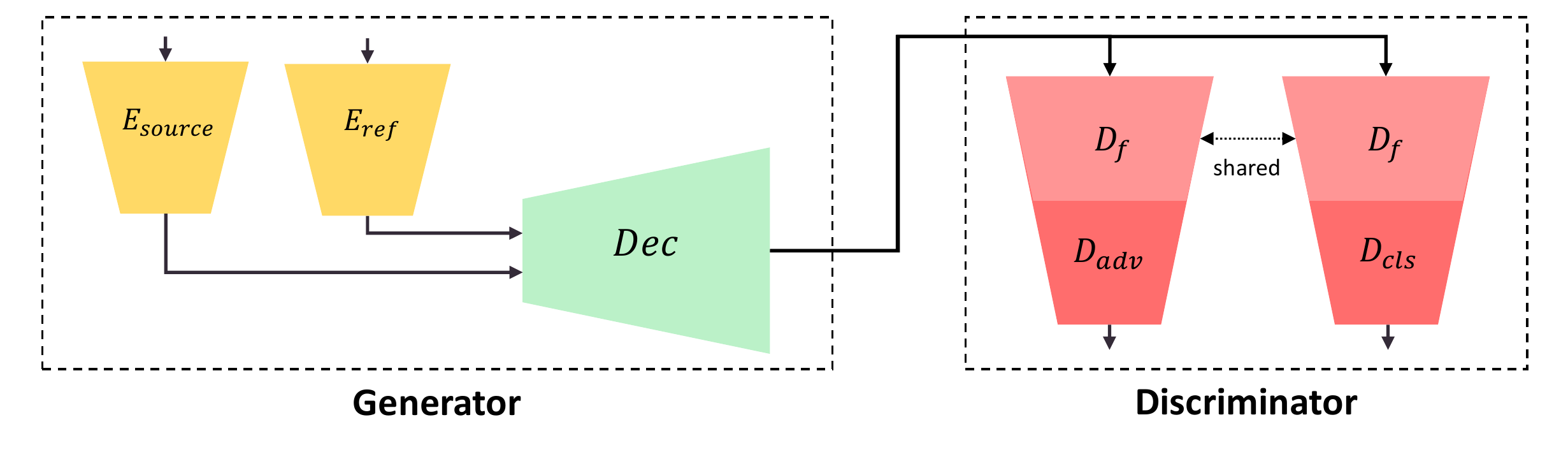}
\end{center}
\caption{An illustration of BalaGAN's architecture.}
\label{fig:architecture}
\end{figure*}

\section{Related Work}

Modern unsupervised image-to-image translation methods use GANs \cite{Goodfellow2014GenerativeAN} to generate plausible images in the target domain, conditioned on images from a source domain. Such methods are unsupervised in the sense that no pairs between the source and target domain are given. 
Some works \cite{CycleGAN2017, NIPS2017_6672, katzir2019crossdomain, GANHopper2020, park2020cut} propose to learn a deterministic generator, which maps each image of the source domain to a corresponding image of the target domain. These works often use a cycle consistency constraint, which enforces the generator to be bijective, thus preventing mode collapse.
With this approach, the amount of possible target images one can generate per input image is often limited. 

Other works \cite{huang2018munit, DRIT_plus} propose to view image-to-image translation as a style transfer problem, where the content is an image from the source domain, and the style is taken from the target domain. The style can be either a random noise from the desired style space or taken from some specific reference image in the target domain. By doing so, the number of possible target images that one can generate significantly increases. These works are multi-modal in the sense that a given image can be translated to multiple images in the target domain. This multi-modality can also be achieved in other approaches \cite{nizan2020breaking}.

While the aforementioned methods require training a generator for each pair of domains, some other works \cite{liu2019few, choi2020starganv2} combine style transfer with a training scheme that results in a single generator that can translate between any pair of domains or styles that appear during training. Moreover, Liu \textit{et al}. \cite{liu2019few} show that their method is capable of translating to styles that were unseen during training as long as the GAN was trained on closely-related styles.

In our work, we adopt the style transfer approach and use the training scheme that enables one generator to translate between multiple pairs of domains. While previous works focus on learning the translation between the desired domains, we also learn translations between \textit{modalities} of the source domain, thus leveraging its richness. This makes our method multi-modal in the sense that it utilizes the modalities of the source domain for the training of the translation task. 
Although the apparent resemblance, the meaning of multi-modal (or cross-modal) in our work is fundamentally different than its meaning in MUNIT, in which multi-modality refers to the ability to translate a given image into multiple images in the target domain. Conversely, in our work, we refer to the latent modalities in the source domain.

Recently, it has been shown that the latent modalities of a dataset can assist in generating images, which belong to that dataset distribution \cite{liu2020diverse, kmodal-stylegan}. The premise of these works is that real-world datasets cannot be well-represented using a uniform latent space, and information about their latent modalities helps to model the data distribution better. In our work, we exploit these modalities to improve the generator by training it to translate between them.

\section{Method}

BalaGAN aims at translating an image between the unpaired, rich source domain $A$, and a data-poor target domain $B$.
\op{As $A$ is rich, it is likely to have a complex distribution, consisting of multiple \textit{modes}.}
To perform the translation, our method receives a source image\op{,} and a reference image from the target domain. 
The source image is translated such that the output image appears to belong to the target domain.
The training of our model consists of two steps: (i) finding $k$ disjoint modalities in the source domain, where each modality is a set of images, denoted by $A_i$; (ii) training a single model to perform cross-translations among all pairs in $(A_1,...,A_k,B)$, see Figure \ref{fig:overview}. \op{By learning all these cross translations, the performance of the network \dc{is significantly improved}, which results in a higher quality of translation from $A$ to $B$, in particular.}

\subsection{Finding Modalities}
To find the modalities of a given domain, we train an encoder that yields a meaningful representation of the style for each image. Then, we cluster the representations of all source domain images, where each cluster represents a single modality.

We train our encoder following Chen \textit{et al.} \cite{chen2020simple}, where contrastive loss is applied on a set of augmented images.
Given a batch of images, we apply two randomly sampled sets of augmentations on each image. Then, we apply the encoder, and attract the result representations of augmented images if both were obtained from the same source image, and repel them otherwise.
Choosing a set of augmentations that distort only content properties of the images, yields representations that are content agnostic and reflecting of the style.
We use the normalized temperature-scaled cross-entropy loss \cite{chen2020simple, NIPS2016_6200, wu2018unsupervised, oord2018representation} to encourage a large cosine similarity between image representations with similar styles.
As the dot product between such representations
is small, spherical $k$-means allows for clustering images by their styles. We denote the clusters by $A_1,...,A_k$, where $k$ is chosen such that $|B| \geq |A|/k$ resulting in modalities which are relatively balanced. Analysis of different values of that $k$ is given in Section \ref{sec:k-analysis}.

\subsection{Translation Network}
Our translation network is a multi-class image-to-image translation network, where the classes $(A_1,...A_k,B)$ are the clusters obtained above. 
The network cross-translates between all the $(k+1)^2$ pairs in $(A_1,...A_k,B)$. The network's architecture and training procedure are built upon FUNIT \cite{liu2019few}.
We train the translation network $G$, and a discriminator $D$ in an adversarial manner. 
A high-level diagram of our architecture is shown in Figure \ref{fig:architecture}.

$G$ consists of source encoder $E_\text{source}$, \op{reference} encoder $E_\text{ref}$, and decoder $F$. Given a source image $x$, and a reference image $y$, the translated image is given by:
\begin{equation}
    x' = G(x,y) = F(E_{\text{source}}(x), E_{\text{ref}}(y)).
\end{equation}
To train $G$, we sample two images from $(A_1,...,A_k,B)$, a source image $x$, and a reference image $y$. The network receives these two images and generates a new image which resembles $x$ on the pixel-wise level, but shares the same mode as $y$.
At test time, we translate images from domain $A$ to domain $B$ by taking a source image from $A$ and a reference image from $B$.
Note that the trained network can translate any image from $A$ without its cluster (modality) label.

Our discriminator consists of two sub-networks, which have shared weights in the initial layers, denoted by $D_f$. 
Each sub-network corresponds to a different task that the discriminator performs.
The first sub-network, denoted by $D_\text{adv}$, aims to solve an adversarial task, in which it classifies each image to one of $(A_1,...,A_k,B)$. That is, $D_\text{adv}(\cdot)$ is a $k+1$-dimensional vector with score for each modality. The translation network aims to confuse the discriminator, that is, given a source image $x$ and a reference image $y$, $G$ aims at making $D_\text{adv}$ predict the modality of $y$ for $G(x,y)$. For such a generated image, $D_\text{adv}$ aims to predict any modality, but the modality of $y$, while for a real image it aims at predicting its correct modality.
\op{The requirement of predicting any modality but the one of the reference image for a generated image is a rather weak requirement, which weakens the discriminator. To strengthen the discriminator, we introduce the additional sub-network, $D_\text{cls}$, which presents a stronger requirement. 
Hence, the shared weights of the two sub-networks learn stronger features.} $D_{\text{cls}}$ is trained to predict the modalities of real images only. 
As shown in previous works, e.g.\cite{Chen_2019_CVPR}, defining additional meaningful task for the discriminator helps the stability of the training, and eventually strengthens the generator. In Section \ref{sec:ablation} we show that this additional sub-network significantly outperforms the FUNIT architecture.

\paragraph{Losses.}
We use a weighted combination of several objectives to train $G$ and $D$.
First, we utilize the Hinge version of the GAN loss for the adversarial loss \cite{Lim2017GeometricG, miyato2018spectral, brock2018large}. It is given by
\begin{align}
    \mathcal{L}_\text{GAN}(D) = E_{x}[\max(0, 1-D_\text{adv}(x)_{m(x)})] + \\
    E_{x,y}[\max(0, 1 + D_\text{adv}(G(x,y))_{m(y)}], \nonumber
\end{align}
\begin{equation*}
    \mathcal{L}_\text{GAN}(G) = -E_{x,y}[D_\text{adv}(G(x,y))_{m(y)}],
\end{equation*}
where $D_\text{adv}(\cdot)_i$ is the $i$-th index in the $k+1$-dimensional vector $D_\text{adv}(\cdot)$ and $m(x)$ is the modality of the image $x$.
To encourage content-preservation of the source image and to help in preventing mode collapse we use a reconstruction loss. It is given by
\begin{equation}
    \mathcal{L}_\text{R}(G) = E_x[||x-G(x,x)||_1].
\end{equation}
Additionally, to encourage the output image to resemble the reference image, we utilize the feature matching loss. It is given by
\begin{equation}
    \mathcal{L}_\text{FM}(G) = E_{x,y}[||D_f(G(x,y)) - D_f(y)||_1].
\end{equation}
For the classification task of the discriminator, we use cross-entropy loss, defined by
\begin{equation}
    \mathcal{L}_\text{CE}(D) = \text{CrossEntropy}(D_\text{cls}(x), \mathbf{1}_{m(x)}),
\end{equation}
where $\mathbf{1}_{m(x)}$ is a one-hot vector that indicates the modality of the image $x$.
Gradient penalty regularization term \cite{Mescheder2018WhichTM} is also utilized, given by
   $ R_1(D) = E_x[||\nabla D_{\text{adv}}(x)||_2^2]$.
The total optimization problem solved by our method is defined by
\begin{equation}
    \min_D \mathcal{L}_\text{GAN}(D) + \lambda_\text{CE} \mathcal{L}_\text{CE}(D) + \lambda_\text{reg}R_1(D),
\end{equation}
\begin{equation*}
    \min_G \mathcal{L}_\text{GAN}(G) + \lambda_\text{R} \mathcal{L}_\text{R}(G) + \lambda_\text{F} \mathcal{L}_\text{FM}(G).    
\end{equation*}

\paragraph{Balanced setting.} \label{method_balanced}
While the main motivation for the \op{cross}-modal translation is for the imbalanced translation setting, our method also shows effectiveness in translation between two balanced domains, $A$ and $B$.
In such a setting, we split both $A$ and $B$ into modalities. Then, instead of defining the classes as $(A_1,...,A_k,B)$, we define the classes to be $(A_1,...,A_{k_s},B_1,...,B_{k_t})$ and train the translation network with all $(k_s + k_t)^2$ pairs.

\section{Evaluation}

\begin{figure*}[h!]
    \setlength{\tabcolsep}{1pt}
    \centering
        \begin{tabu}{c c@{\hspace{2.5pt}} |[0.5pt]@{\hspace{2.5pt}} c c c c @{\hspace{2.5pt}}|[0.5pt]@{\hspace{2.5pt}} c c c c}
            \multicolumn{2}{c}{} & 1000 & 500 & 250 & \multicolumn{1}{c}{125} & 1000 & 500 & 250 & 125
            \tabularnewline
            \includegraphics[height=0.0935\textwidth]{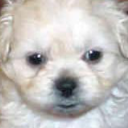} &
            \includegraphics[height=0.0935\textwidth]{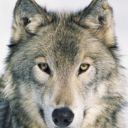} &
            \includegraphics[height=0.0935\textwidth]{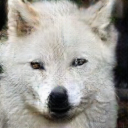} &
            \includegraphics[height=0.0935\textwidth]{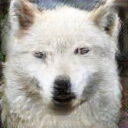} &
            \includegraphics[height=0.0935\textwidth]{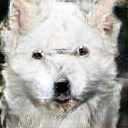} &
            \includegraphics[height=0.0935\textwidth]{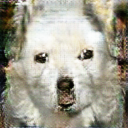} &
            \includegraphics[height=0.0935\textwidth]{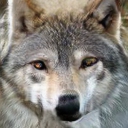} &
            \includegraphics[height=0.0935\textwidth]{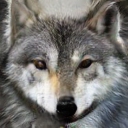} &
            \includegraphics[height=0.0935\textwidth]{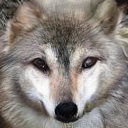} &
            \includegraphics[height=0.0935\textwidth]{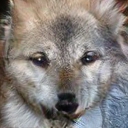} 
            \tabularnewline
            \includegraphics[height=0.0935\textwidth]{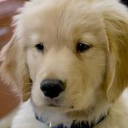} &
            \includegraphics[height=0.0935\textwidth]{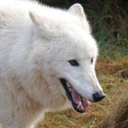} &
            \includegraphics[height=0.0935\textwidth]{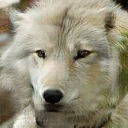} &
            \includegraphics[height=0.0935\textwidth]{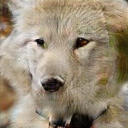} &
            \includegraphics[height=0.0935\textwidth]{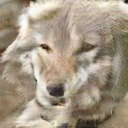} &
            \includegraphics[height=0.0935\textwidth]{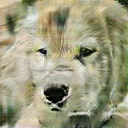} &
            \includegraphics[height=0.0935\textwidth]{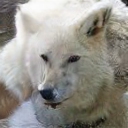} &
            \includegraphics[height=0.0935\textwidth]{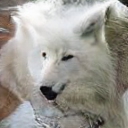} &
            \includegraphics[height=0.0935\textwidth]{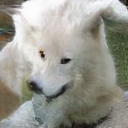} &
            \includegraphics[height=0.0935\textwidth]{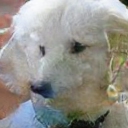}
            \tabularnewline
            \includegraphics[height=0.0935\textwidth]{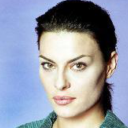} &
            \includegraphics[height=0.0935\textwidth]{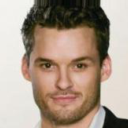} &
            \includegraphics[height=0.0935\textwidth]{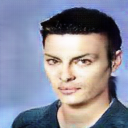} &
            \includegraphics[height=0.0935\textwidth]{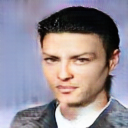} &
            \includegraphics[height=0.0935\textwidth]{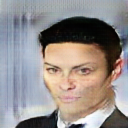} &
            \includegraphics[height=0.0935\textwidth]{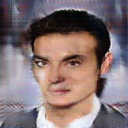} &
            \includegraphics[height=0.0935\textwidth]{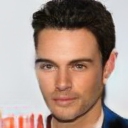} &
            \includegraphics[height=0.0935\textwidth]{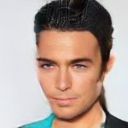} &
            \includegraphics[height=0.0935\textwidth]{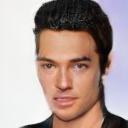} &
            \includegraphics[height=0.0935\textwidth]{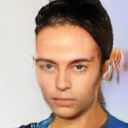} 
            \tabularnewline
            \includegraphics[height=0.0935\textwidth]{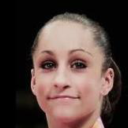} &
            \includegraphics[height=0.0935\textwidth]{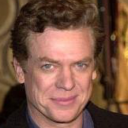} &
            \includegraphics[height=0.0935\textwidth]{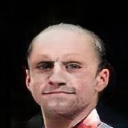} &
            \includegraphics[height=0.0935\textwidth]{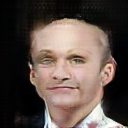} &
            \includegraphics[height=0.0935\textwidth]{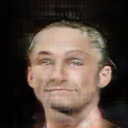} &
            \includegraphics[height=0.0935\textwidth]{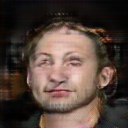} &
            \includegraphics[height=0.0935\textwidth]{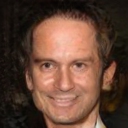} &
            \includegraphics[height=0.0935\textwidth]{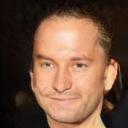} &
            \includegraphics[height=0.0935\textwidth]{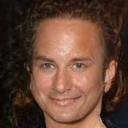} &
            \includegraphics[height=0.0935\textwidth]{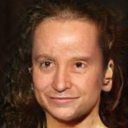} 
            \tabularnewline
            \includegraphics[height=0.0935\textwidth]{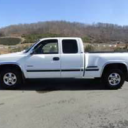} &
            \includegraphics[height=0.0935\textwidth]{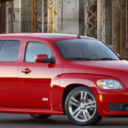} & 
            \includegraphics[height=0.0935\textwidth]{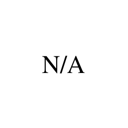} &
            \includegraphics[height=0.0935\textwidth]{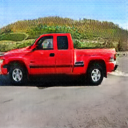} &
            \includegraphics[height=0.0935\textwidth]{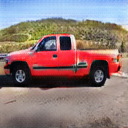} &
            \includegraphics[height=0.0935\textwidth]{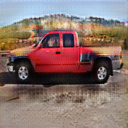} &
            \includegraphics[height=0.0935\textwidth]{img/na_small.png} &
            \includegraphics[height=0.0935\textwidth]{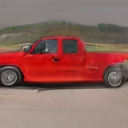} &
            \includegraphics[height=0.0935\textwidth]{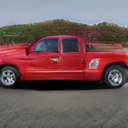} &
            \includegraphics[height=0.0935\textwidth]{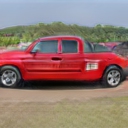}
            \tabularnewline
            \includegraphics[height=0.0935\textwidth]{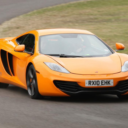} &
            \includegraphics[height=0.0935\textwidth]{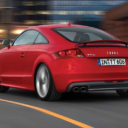} & 
            \includegraphics[height=0.0935\textwidth]{img/na_small.png} &
            \includegraphics[height=0.0935\textwidth]{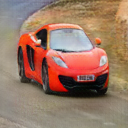} &
            \includegraphics[height=0.0935\textwidth]{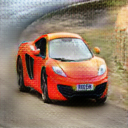} &
            \includegraphics[height=0.0935\textwidth]{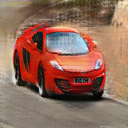} &
            \includegraphics[height=0.0935\textwidth]{img/na_small.png} &
            \includegraphics[height=0.0935\textwidth]{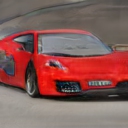} &
            \includegraphics[height=0.0935\textwidth]{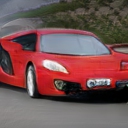} &
            \includegraphics[height=0.0935\textwidth]{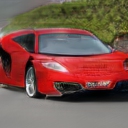}
            \tabularnewline
            \includegraphics[height=0.0935\textwidth]{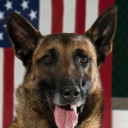} &
            \includegraphics[height=0.0935\textwidth]{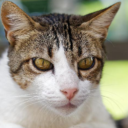} &
            \includegraphics[height=0.0935\textwidth]{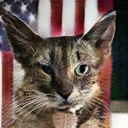} &
            \includegraphics[height=0.0935\textwidth]{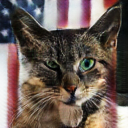} &
            \includegraphics[height=0.0935\textwidth]{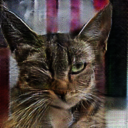} &
            \includegraphics[height=0.0935\textwidth]{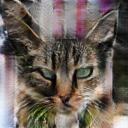} &
            \includegraphics[height=0.0935\textwidth]{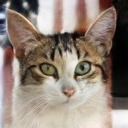} &
            \includegraphics[height=0.0935\textwidth]{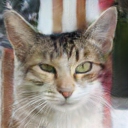} &
            \includegraphics[height=0.0935\textwidth]{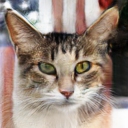} &
            \includegraphics[height=0.0935\textwidth]{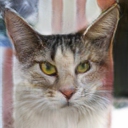} 
            \tabularnewline
            \includegraphics[height=0.0935\textwidth]{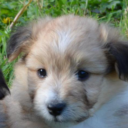} &
            \includegraphics[height=0.0935\textwidth]{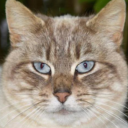} &
            \includegraphics[height=0.0935\textwidth]{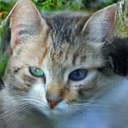} &
            \includegraphics[height=0.0935\textwidth]{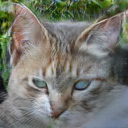} &
            \includegraphics[height=0.0935\textwidth]{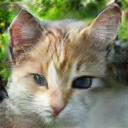} &
            \includegraphics[height=0.0935\textwidth]{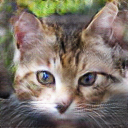} &
            \includegraphics[height=0.0935\textwidth]{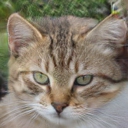} &
            \includegraphics[height=0.0935\textwidth]{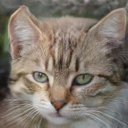} &
            \includegraphics[height=0.0935\textwidth]{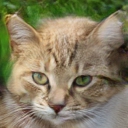} &
            \includegraphics[height=0.0935\textwidth]{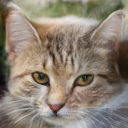} 
            \tabularnewline
            Source & \multicolumn{1}{c}{Ref} & \multicolumn{4}{c}{CycleGAN} & \multicolumn{4}{c}{BalaGAN}
        \end{tabu}
        \caption{Applying CycleGAN and BalaGAN on the dog$\rightarrow$wolf, woman$\rightarrow$men, car$\rightarrow$red-car, and dog$\rightarrow$cat translation tasks, by training with decreasing number of images in the target domain. 
        The numbers above the table indicate the number of target domain images that were used for training. \op{For the cars, we have less than 1000 red car images overall, and thus training with 1000 images in the target domain is not applicable.}}
        \label{fig:unbalanced_dogs_wolfs}
\end{figure*}

We evaluate our cross-modal translation method in a series of experiments.  
We first show the effectiveness of our method in the imbalanced setting, by evaluating its performance when \op{training it with a} decreasing the number of images in the target domain.
Next, we explore the influence of the number of modalities, $k$, on the result. Then, we show that our method can also be effective in the balanced setting. Finally, we perform an ablation study to compare our architecture with other alternative architectures and study the importance of finding effective modalities.
To evaluate the results, we show a variety of visual examples, use the FID \cite{heusel2017gans} measurement, and perform a human perceptual study to validate the quality of the results obtained by our method compared to results of other leading methods.

\paragraph{Datasets.} We use the CelebA dataset \cite{liu2015faceattributes} and set the source and target domains to consist of 10,000 and 1000 images of women and men, respectively.
We additionally use the Stanford Cars Dataset \cite{KrauseStarkDengFei-Fei_3DRR2013}, and translate a range of different colored cars to red cars. There, the training set consists of 7500 non-red cars, and 500 red cars. 
From the AFHQ dataset \cite{choi2020starganv2} we take all the 4739 images of dogs as the source domain, and all the 5153 images of cats as the target domain.
Furthermore, we use the Animal Face Dataset (AFD) \cite{liu2019few} and set the source domain to be a mix of 16 breeds of dogs and the target domain to be a mix of three breeds of wolves. Our training set consists of 10,000 dog images and 1000 wolf images. 
It should be noted that among the above, the Animal Face Dataset is the most challenging due to the wide range of poses and image quality.

\subsection{Effectiveness in the Imbalanced Setting} 

We compare our approach with other methods: CycleGAN \cite{CycleGAN2017}, CUT \cite{park2020cut}, U-GAT-IT \cite{Kim2020UGATITUG}, MUNIT \cite{huang2018munit}, StarGAN2 \cite{choi2020starganv2}. We first train a number of methods on the AFD dataset. For our method, we used 40 modalities to train the translation network. Quantitative results are presented in Table \ref{tab:unbalanced_fid_all}.

\begin{table}[h]
    \centering
    \begin{tabular}{|c|c|c|c|c|c|}
    \hline
    \cite{CycleGAN2017} & \cite{park2020cut} & \cite{Kim2020UGATITUG} & \cite{huang2018munit} & \cite{choi2020starganv2} & \textbf{Ours} \\
    \hline \hline
    77.8 & 108.64 & 97.16 & 83.38 & 211.77 & \textbf{60.88}\\
    \hline
    \end{tabular}
    \caption{FID ($\downarrow$) \op{results of CycleGAN, CUT, U-GAT-IT, MUNIT, StarGAN2, and BalaGAN (marked by their references) applied on AFD, translating dogs to wolves in an imbalanced setting. For BalaGAN we use 40 modalities.}}
    \label{tab:unbalanced_fid_all}
\end{table}

\begin{table}
    \centering
\begin{tabular}{|c | c | c|}
\hline
    \textbf{Task} & \textbf{CycleGAN} & \textbf{BalaGAN} \\
    \hline
    \hline
    dogs $\rightarrow$ wolfs & 16.7 & \textbf{83.3} \\
    women $\rightarrow$ men & 33.6 & \textbf{66.4} \\
    \hline
\end{tabular} 
    \caption{Percentage of users that chose the corresponding image as the preferred one in imbalanced setting.}
    \label{tab:user_unbalanced}
\end{table}

\setlength\tabcolsep{1pt}
\begin{table*}[h]
    \centering
    \begin{tabular}{|c||c|c|c|c||c|c|c|c||c|c||c|c|}
    \hline
        \multirow{2}{*}{$\mathbf{|B|}$} &\multicolumn{4}{c||}{\textbf{dogs$\rightarrow$wolves}} & \multicolumn{4}{c|}{\textbf{women$\rightarrow$men}} & \multicolumn{2}{c|}{\textbf{car$\rightarrow$red-car}} & \multicolumn{2}{c|}{\textbf{dog$\rightarrow$cat}}\\
        \cline{2-13}
         & \textbf{CycleGAN} & \textbf{CUT} & \textbf{MUNIT} & \textbf{Ours} & \textbf{CycleGAN} & \textbf{CUT} & \textbf{MUNIT} & \textbf{Ours} & \textbf{CycleGAN} & \textbf{Ours} & \textbf{CycleGAN} & \textbf{Ours}\\
        \hline \hline
        1000 & 77.80 & 108.64 & 83.38 & \textbf{60.88} & \textbf{28.33} & 55.04 & 42.35 & 33.42 & N/A & N/A & 29.64 & \textbf{26.21}\\
        500 & 99.80 & 166.36 & 103.07 & \textbf{72.46} & \textbf{38.59} & 61.08 & 47.51 & 39.95 & \textbf{33.38} & 37.02 & 30.71 & \textbf{28.71}\\
        250 & 136.00 & 225.35 & 123.88 & \textbf{102.35} & 54.95 & 82.26 & 53.81 &  \textbf{38.99} & 40.46 & 40.38 & 38.61 & \textbf{37.89}\\
        125 & 202.61 & 226.97 & 162.97 & \textbf{157.67}& 155.60 & 274.53 & 58.48 & \textbf{49.42} & 51.17 & \textbf{40.25} & 49.58 & \textbf{45.14}\\
        \hline
    \end{tabular}
    \caption{FID results ($\downarrow$) applied on AFD, CelebA, Stanford Cars, and AFHQ datasets in the imbalanced setting. $|B|$ denotes the number of images in the target domain that were used during training. }
    \label{tab:unbalanced_fid}
\end{table*}

For the above leading methods, we perform additional experiments over multiple datasets to show the effect of decreasing the number of training images in the target domain. Quantitative results over AFD, CelebA, Stanford Cars, and AFHQ are presented in Table \ref{tab:unbalanced_fid}. As can be seen, CycleGAN and BalaGAN are the leading methods, and the image quality produced by BalaGAN is more stable as the size of the target domain decreases. Visual results are shown in Figure \ref{fig:unbalanced_dogs_wolfs} for these two methods, and in the supplementary material for the other methods.

We further compare BalaGAN and CycleGAN through a human perceptual study, in which each user was asked to select the preferred image between images generated by these two methods. The images were generated by models that were trained using 1000 target domain images.
50 users participated in the survey, each answered ten random questions out of a pool of 200 questions for each dataset. As can be seen in Table \ref{tab:user_unbalanced}, BalaGAN outperforms CycleGAN on both datasets even though CycleGAN achieves lower FID for the women$\rightarrow$men translation task.

\subsection{Influence of \op{Modalities Number}} \label{sec:k-analysis}
\begin{figure}
    \centering
    \includegraphics[width=0.98\textwidth]{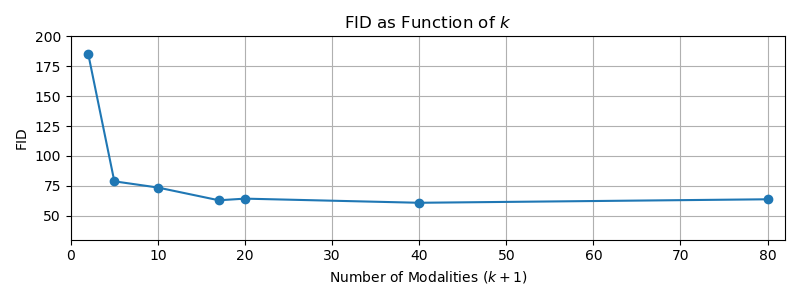}
    \caption{FID ($\downarrow$) of our method applied on AFD in an imbalanced setting. The number of modalities is $k + 1$.}
    \label{fig:increasing_k_graph}
\end{figure}

\begin{figure}
		\setlength{\tabcolsep}{1pt}
		\centering
			\begin{tabular}{c c c c c c c}
				\includegraphics[height=0.13\textwidth]{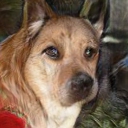} &
				\includegraphics[height=0.13\textwidth]{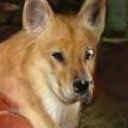} &
				\includegraphics[height=0.13\textwidth]{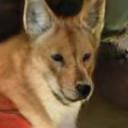} &
				\includegraphics[height=0.13\textwidth]{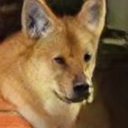} &
				\includegraphics[height=0.13\textwidth]{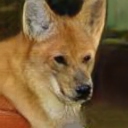} &            
				\includegraphics[height=0.13\textwidth]{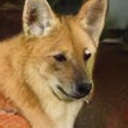} &
				\includegraphics[height=0.13\textwidth]{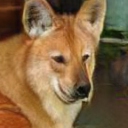}
				\tabularnewline
				\includegraphics[height=0.13\textwidth]{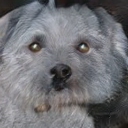} &
				\includegraphics[height=0.13\textwidth]{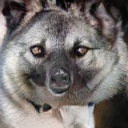} &
				\includegraphics[height=0.13\textwidth]{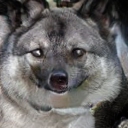} &
				\includegraphics[height=0.13\textwidth]{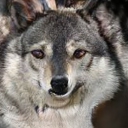} &
				\includegraphics[height=0.13\textwidth]{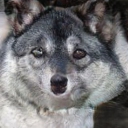} &            
				\includegraphics[height=0.13\textwidth]{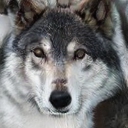} &
				\includegraphics[height=0.13\textwidth]{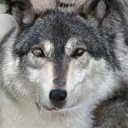}
				\tabularnewline
				\includegraphics[height=0.13\textwidth]{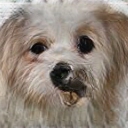} &
				\includegraphics[height=0.13\textwidth]{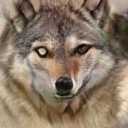} &
				\includegraphics[height=0.13\textwidth]{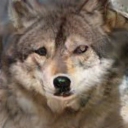} &
				\includegraphics[height=0.13\textwidth]{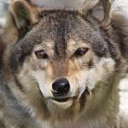} &
				\includegraphics[height=0.13\textwidth]{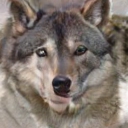} &            
				\includegraphics[height=0.13\textwidth]{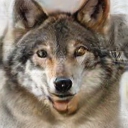} &
				\includegraphics[height=0.13\textwidth]{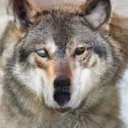}
				\tabularnewline
				$2$ & $5$ & $10$ & $17$ & $20$ & $40$ & $80$
			\end{tabular}
			\caption{Results of BalaGAN applied with varying values of $k$. Below each column we specify the number of modalities that the translation network was trained on, that is $k+1$.}
			\label{fig:increasing_k}
	\end{figure}

The number of modalities that our translation network is trained on, $k+1$, is an important factor for the success of our method. For $k = 1$, our method is reduced to the common setting of image-to-image translation, and as we increase $k$, our network is enforced to train and learn more translation tasks, resulting in more accurate translation. Here we show that the value of $k$ influences the quality of the generated images. Visual results that were obtained on the dog$\rightarrow$wolf translation task are shown in Figure \ref{fig:increasing_k} and quantitative results are provided in Figure \ref{fig:increasing_k_graph}. 
As can be seen, as $k$ increases, FID decreases, i.e., the images quality is improved. Note, however, that once $k$ goes beyond 16, the number of dog breeds, the improvement of the results is rather moderate.

\subsection{Effectiveness in the Balanced Setting} \label{balanced-results}
\begin{figure*}[h]
    \setlength{\tabcolsep}{1pt}
    \centering
        \begin{tabu}{c c @{\hspace{2.5pt}}|[0.5pt]@{\hspace{2.5pt}} c c c c c c c c}
            \includegraphics[height=0.095\textwidth]{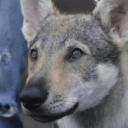} &
            \includegraphics[height=0.095\textwidth]{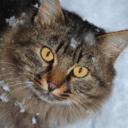} &
            \includegraphics[height=0.095\textwidth]{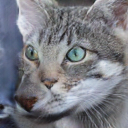} &
            \includegraphics[height=0.095\textwidth]{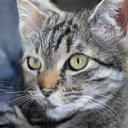} &
            \includegraphics[height=0.095\textwidth]{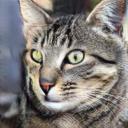} &
            \includegraphics[height=0.095\textwidth]{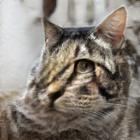} &
            \includegraphics[height=0.095\textwidth]{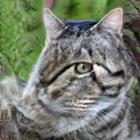} &
            \includegraphics[height=0.095\textwidth]{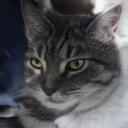} &
            \includegraphics[height=0.095\textwidth]{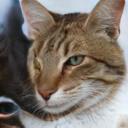} &
            \includegraphics[height=0.095\textwidth]{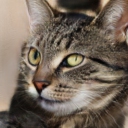}
            \tabularnewline
            \includegraphics[height=0.095\textwidth]{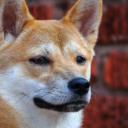} &
            \includegraphics[height=0.095\textwidth]{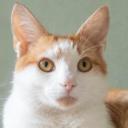} &
            \includegraphics[height=0.095\textwidth]{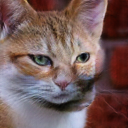} &
            \includegraphics[height=0.095\textwidth]{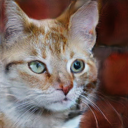} &
            \includegraphics[height=0.095\textwidth]{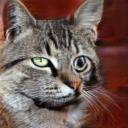} &
            \includegraphics[height=0.095\textwidth]{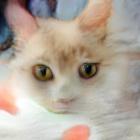} &
            \includegraphics[height=0.095\textwidth]{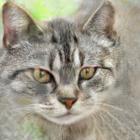} &
            \includegraphics[height=0.095\textwidth]{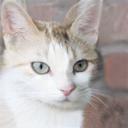} &
            \includegraphics[height=0.095\textwidth]{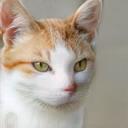} &
            \includegraphics[height=0.095\textwidth]{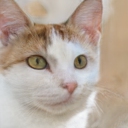}
            \tabularnewline
            \includegraphics[height=0.095\textwidth]{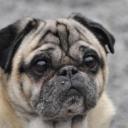} &
            \includegraphics[height=0.095\textwidth]{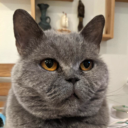} &
            \includegraphics[height=0.095\textwidth]{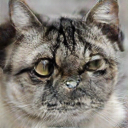} &
            \includegraphics[height=0.095\textwidth]{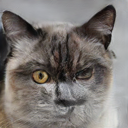} &            
            \includegraphics[height=0.095\textwidth]{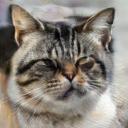} &
            \includegraphics[height=0.095\textwidth]{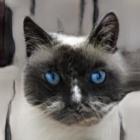} &
            \includegraphics[height=0.095\textwidth]{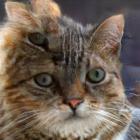} &
            \includegraphics[height=0.095\textwidth]{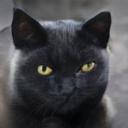} &
            \includegraphics[height=0.095\textwidth]{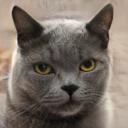} &
            \includegraphics[height=0.095\textwidth]{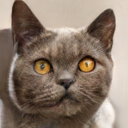}
            \tabularnewline
            \begin{small} Source \end{small}&
            \multicolumn{1}{c}{\begin{small} Ref \end{small}}&
            \begin{small}CycleGAN \end{small}&
            \begin{small}CUT \end{small} &
            \begin{small} Hopper \end{small} &
            \begin{small}MUNIT$^{r}$ \end{small}&
            \begin{small}MUNIT$^{n}$ \end{small}&
            \begin{small}StarGAN$^1$ \end{small}&
            \begin{small}StarGAN$^{30}$ \end{small}&
            \begin{small}BalaGAN$^{30}$\end{small}
        \end{tabu}
        \caption{Various methods applied on AFHQ dataset, which is balanced, to translate dogs to cats. Notations for MUNIT and StarGAN are explained in \ref{balanced-results}. Additional results are shown in the supplementary material.}
        \label{fig:balanced}
    \end{figure*}

Here we present results on a balanced dataset. We choose the AFHQ dataset, translating dogs to cats. 
We train BalaGAN using latent modalities extracted
in both the source and target domain. For this dataset, we extracted 30 modalities in each domain.
We compare our method with five strong baseline methods: CycleGAN \cite{CycleGAN2017}, CUT \cite{park2020cut}, GANHopper \cite{GANHopper2020}, MUNIT \cite{huang2018munit}, and StarGAN2 \cite{choi2020starganv2}.
For MUNIT, we show results when the style is taken from a reference image (denoted by MUNIT$^r$), and from a random noise vector (denoted by MUNIT$^n$).
We denote the StarGAN2 that is trained on the two domains as StarGAN2$^1$, and StarGAN2 that is trained to translate between each pair of the 60 modalities that we find as StarGAN2$^{30}$.
Figure \ref{fig:balanced} shows a random sample of results from this comparison, and in Table \ref{tab:balanced} we present a quantitative comparison. 
As can be observed, our method outperforms other methods both visually and quantitatively.

\begin{table}[h]
    \centering
    \begin{tabular}{|c|c|c|c|c|c|c|c|}
    \hline
    \cite{CycleGAN2017} & \cite{park2020cut} & \cite{GANHopper2020} & \cite{huang2018munit}$^r$ & \cite{huang2018munit}$^n$ & \cite{choi2020starganv2}$^1$ & \cite{choi2020starganv2}$^{30}$ & \textbf{Ours} \\
    \hline
    \hline
    29.98 & 27.37 & 33.79 & 35.80 & 27.11 & 29.56 & 25.89 & \textbf{19.21 } \\
    \hline
    \end{tabular}
    \caption{FID ($\downarrow$) results of applying \op{CycleGAN, CUT, GANHopper, MUNIT$^r$, MUNIT$^n$, StarGAN$^1$, StarGAN2$^{30}$, and BalaGAN} over AFHQ dataset.}
    \label{tab:balanced}
\end{table}

As the leading methods according to the FID measure are BalaGAN and StarGAN2, we further compared them through a human perceptual study.
Similarly to the imbalanced user study, each user answered 10 random questions out of a pool of 200 questions. Here, the user was asked to choose between images of BalaGAN, StarGAN2$^{30}$ and StarGAN2$^1$. As observed in Table \ref{tab:user_balanced}, most users chose images of BalaGAN, where the scores of StarGAN2$^{30}$, and StarGAN2$^1$ are similar.

\subsection{Diversity and Ablation Study} \label{sec:ablation}

\begin{figure}
    
            \setlength{\tabcolsep}{1pt}
    \setlength{\extrarowsep}{-1pt}
    \centering
        \begin{tabu}{c@{\hspace{2pt}} |[0.5pt]@{\hspace{2pt}} c c c c}
            Ref \hspace{7pt}\rotatebox[origin=l]{90}{Source} &
            \includegraphics[height=0.18\textwidth]{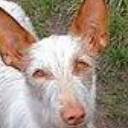} &
            \includegraphics[height=0.18\textwidth]{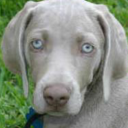} &
            \includegraphics[height=0.18\textwidth]{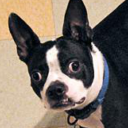} &
            \includegraphics[height=0.18\textwidth]{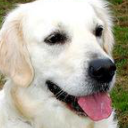} \\
            \Xhline{1\arrayrulewidth}
            \addstackgap[2.5pt]{\includegraphics[height=0.18\textwidth]{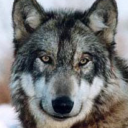}} &
            \includegraphics[height=0.18\textwidth]{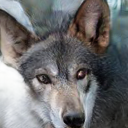} &
            \includegraphics[height=0.18\textwidth]{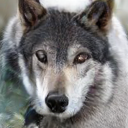} &
            \includegraphics[height=0.18\textwidth]{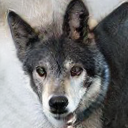} &
            \includegraphics[height=0.18\textwidth]{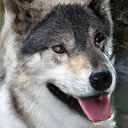} \\
            \includegraphics[height=0.18\textwidth]{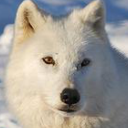} &
            \includegraphics[height=0.18\textwidth]{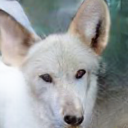} &
            \includegraphics[height=0.18\textwidth]{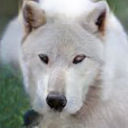} &
            \includegraphics[height=0.18\textwidth]{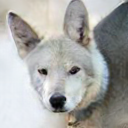} &
            \includegraphics[height=0.18\textwidth]{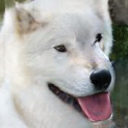} \\
            \includegraphics[height=0.18\textwidth]{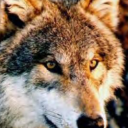} &
            \includegraphics[height=0.18\textwidth]{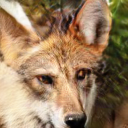} &
            \includegraphics[height=0.18\textwidth]{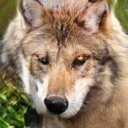} &
            \includegraphics[height=0.18\textwidth]{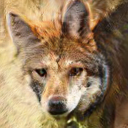} &
            \includegraphics[height=0.18\textwidth]{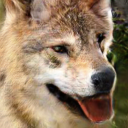}
        \end{tabu}
        \caption{BalaGAN on the AFD trained on 1000 wolves using 40 modalities.}
        \label{fig:matrix-wolfs-1}
\end{figure}

\begin{figure}
    \centering
    \setlength{\tabcolsep}{1.1pt}
    		\begin{tabu}{c c c c c c}
            \includegraphics[height=0.15\textwidth]{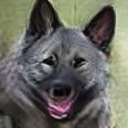} &
            \includegraphics[height=0.15\textwidth]{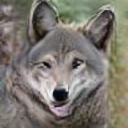} &
            \includegraphics[height=0.15\textwidth]{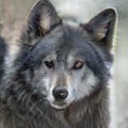} &
            \includegraphics[height=0.15\textwidth]{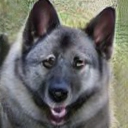} &
            \includegraphics[height=0.15\textwidth]{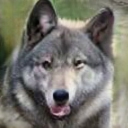} &
            \includegraphics[height=0.15\textwidth]{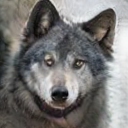} \\ 
            \texttt{\textbf{C}}/Ours & \texttt{\textbf{B}}/Ours & \texttt{\textbf{A}}/Ours & \texttt{\textbf{C}}/GT & \texttt{\textbf{B}}/GT & \texttt{\textbf{A}}/GT
            
        \end{tabu}
    \caption{\op{Visual results of ablation study. \texttt{\textbf{A}}, \texttt{\textbf{B}}, and \texttt{\textbf{C}} correspond to the architecture versions of the translation network as explained in \ref{sec:ablation}. ``Ours'' corresponds to the BalaGAN domain decomposition, and ``GT'' corresponds to the ground-truth labels of the AFD.}}
    \label{fig:ablation}
\end{figure}

\begin{table}[]
    \centering
		\begin{tabular}{|c|c|c|c|c|}
		\hline
		& \textbf{AFD GT} & \textbf{mVE} & \textbf{Ours} \\
		\hline
		\hline
		\makecell[l]{\texttt{\textbf{A}}} & \textbf{49.89} & \textbf{151.27} & \textbf{62.97}\\
		\makecell[l]{\texttt{\textbf{B}}} & 72.87 & 190.33 & 100.67\\
		\makecell[l]{\texttt{\textbf{C}}} & 168.94 & 202.90 & 187.76\\
		\hline
		\end{tabular}
			\caption{FID($\downarrow$) of our ablation study applied on the dog$\rightarrow$wolf translation task with $k = 16$ which is the number of dogs' breeds. Notations are explained in \ref{sec:ablation}}
		\label{tab:ablation} 
\end{table}

\begin{table}[]
    \centering
\begin{tabular}{|c|c|c|}
\hline
     \textbf{StarGAN2$^{30}$} & \textbf{StarGAN2$^1$} & \textbf{BalaGAN}\\
     \hline
     \hline
     27.4 & 28.7 & \textbf{43.9}  \\
     \hline
\end{tabular}
    \caption{Users preferences for the AFHQ dataset in a balanced setting. We present the percentage of users that chose the corresponding image as the preferred one.}
    \label{tab:user_balanced}
\end{table}

The diversity of generated images that is achieved by our method, is shown in Figure \ref{fig:matrix-wolfs-1} (see additional results in the supplementary material).
We also perform an ablation study, \op{in which we examine the results obtained by various combinations of modified translation network and source domain decomposition.}
For the ablation of the translation network, let \texttt{\textbf{A}} denote our BalaGAN method, then (i) in \texttt{\textbf{B}} we removed the $D_\text{cls}$ loss, and (ii) in \texttt{\textbf{C}}, we additionally do not use the target domain images during training. Note, that the setting in \texttt{\textbf{C}} degenerates into FUNIT \cite{liu2019few}. 
For the ablation of the source's decomposition, let AFD GT denote the dogs' breeds ground truth class labels, \op{that is, this decomposition is a natural one, and requires supervision. 
Let mVE denotes the modalities that are obtained by clustering representations achieved by a variational autoencoder.} The results presented in Table \ref{tab:ablation} and Figure \ref{fig:ablation} show that $D_\text{cls}$ significantly improves FUNIT, even in a transductive setting.

\op{In the following, we show that the core idea of training a translation network with cross-modalities, can contribute to different translation network architectures, other than the one that we have presented above. However, unlike CycleGAN and MUNIT, the translation network needs to support translation between multiple domains or classes, like StarGAN2.
We demonstrate it on StarGAN2 architecture and compare} the results of two variations of it, one is trained on the source and target domains, and the other on our modalities, denoted by StarGAN2$^{1}$ and StarGAN2$^{30}$, respectively. The results are shown in Figure \ref{fig:balanced} and Table \ref{tab:balanced}.
As one can see, training StarGAN2 to translate between modalities improves the network's ability to translate between the two domains.
Therefore, we conclude that the benefit of training on modalities is not specific to our architecture, and can be utilized by other multi-class image-to-image translation methods.

\section{Conclusion}

We have presented an image-to-image translation technique that leverages latent modes in the source and target domains. The technique was designed to alleviate the problems associated with the imbalanced setting, where the target domain is poor. The key idea is to convert the imbalanced setting to a \op{more }balanced one, where the network is trained to translate between all pairs of modes, including the target one. We have shown that \op{this} setting leads to better translation than strong baselines. We further showed that analyzing and translating at the mode-level, can benefit also in a balanced setting, where both the source and target domains are split and the translator is trained on all pairs.
In the future, we would like to use our technique to re-balance training sets and show that downstream applications, like object classification and detection, can benefit from the re-balancing operation.
We believe our work to be a step in the direction of analyzing domain distributions and learning their latent modes, and would like to reason and apply this idea on a wider range of problems beyond image-to-image translation.

\clearpage
{\small
\bibliographystyle{ieee_fullname}
\bibliography{egbib}

\begin{thebibliography}{10}\itemsep=-1pt

\bibitem{brock2018large}
Andrew Brock, Jeff Donahue, and Karen Simonyan.
\newblock Large scale {GAN} training for high fidelity natural image synthesis.
\newblock In {\em International Conference on Learning Representations}, 2019.

\bibitem{chen2020simple}
Ting Chen, Simon Kornblith, Mohammad Norouzi, and Geoffrey Hinton.
\newblock A simple framework for contrastive learning of visual
  representations.
\newblock {\em arXiv preprint arXiv:2002.05709}, 2020.

\bibitem{Chen_2019_CVPR}
Ting Chen, Xiaohua Zhai, Marvin Ritter, Mario Lucic, and Neil Houlsby.
\newblock Self-supervised gans via auxiliary rotation loss.
\newblock In {\em Proceedings of the IEEE/CVF Conference on Computer Vision and
  Pattern Recognition (CVPR)}, June 2019.

\bibitem{choi2020starganv2}
Yunjey Choi, Youngjung Uh, Jaejun Yoo, and Jung-Woo Ha.
\newblock Stargan v2: Diverse image synthesis for multiple domains.
\newblock In {\em Proceedings of the IEEE Conference on Computer Vision and
  Pattern Recognition}, 2020.

\bibitem{imagenet_cvpr09}
J. Deng, W. Dong, R. Socher, L.-J. Li, K. Li, and L. Fei-Fei.
\newblock {ImageNet: A Large-Scale Hierarchical Image Database}.
\newblock In {\em CVPR09}, 2009.

\bibitem{Goodfellow2014GenerativeAN}
Ian~J. Goodfellow, Jean Pouget-Abadie, Mehdi Mirza, Bing Xu, David
  Warde-Farley, Sherjil Ozair, Aaron Courville, and Yoshua Bengio.
\newblock Generative adversarial nets.
\newblock In {\em Proceedings of the 27th International Conference on Neural
  Information Processing Systems - Volume 2}, NIPS’14, page 2672–2680,
  Cambridge, MA, USA, 2014. MIT Press.

\bibitem{heusel2017gans}
Martin Heusel, Hubert Ramsauer, Thomas Unterthiner, Bernhard Nessler, and Sepp
  Hochreiter.
\newblock Gans trained by a two time-scale update rule converge to a local nash
  equilibrium.
\newblock In {\em Advances in neural information processing systems}, pages
  6626--6637, 2017.

\bibitem{huang2018munit}
Xun Huang, Ming-Yu Liu, Serge Belongie, and Jan Kautz.
\newblock Multimodal unsupervised image-to-image translation.
\newblock In {\em ECCV}, 2018.

\bibitem{katzir2019crossdomain}
Oren Katzir, Dani Lischinski, and Daniel Cohen-Or.
\newblock Cross-domain cascaded deep feature translation.
\newblock In {\em ECCV}, 2020.

\bibitem{Kim2020UGATITUG}
Junho Kim, Minjae Kim, Hyeonwoo Kang, and Kwang~Hee Lee.
\newblock {U-GAT-IT}: Unsupervised generative attentional networks with
  adaptive layer-instance normalization for image-to-image translation.
\newblock In {\em International Conference on Learning Representations}, 2020.

\bibitem{KrauseStarkDengFei-Fei_3DRR2013}
Jonathan Krause, Michael Stark, Jia Deng, and Li Fei-Fei.
\newblock 3d object representations for fine-grained categorization.
\newblock In {\em 4th International IEEE Workshop on 3D Representation and
  Recognition (3dRR-13)}, Sydney, Australia, 2013.

\bibitem{DRIT_plus}
Hsin-Ying Lee, Hung-Yu Tseng, Qi Mao, Jia-Bin Huang, Yu-Ding Lu, Maneesh~Kumar
  Singh, and Ming-Hsuan Yang.
\newblock Drit++: Diverse image-to-image translation viadisentangled
  representations.
\newblock {\em arXiv preprint arXiv:1905.01270}, 2019.

\bibitem{Lim2017GeometricG}
J.~H. Lim and J.~C. Ye.
\newblock Geometric gan.
\newblock {\em ArXiv}, abs/1705.02894, 2017.

\bibitem{GANHopper2020}
Wallace Lira, Johannes Merz, Daniel Ritchie, Daniel Cohen-Or, and Hao Zhang.
\newblock Ganhopper: Multi-hop gan for unsupervised image-to-image translation,
  2020.

\bibitem{NIPS2017_6672}
Ming-Yu Liu, Thomas Breuel, and Jan Kautz.
\newblock Unsupervised image-to-image translation networks.
\newblock In I. Guyon, U.~V. Luxburg, S. Bengio, H. Wallach, R. Fergus, S.
  Vishwanathan, and R. Garnett, editors, {\em Advances in Neural Information
  Processing Systems 30}, pages 700--708. Curran Associates, Inc., 2017.

\bibitem{liu2019few}
Ming-Yu Liu, Xun Huang, Arun Mallya, Tero Karras, Timo Aila, Jaakko Lehtinen,
  and Jan Kautz.
\newblock Few-shot unsupervised image-to-image translation.
\newblock In {\em IEEE International Conference on Computer Vision (ICCV)},
  2019.

\bibitem{liu2020diverse}
Steven Liu, Tongzhou Wang, David Bau, Jun-Yan Zhu, and Antonio Torralba.
\newblock Diverse image generation via self-conditioned gans.
\newblock In {\em Proceedings of the IEEE/CVF Conference on Computer Vision and
  Pattern Recognition}, pages 14286--14295, 2020.

\bibitem{liu2015faceattributes}
Ziwei Liu, Ping Luo, Xiaogang Wang, and Xiaoou Tang.
\newblock Deep learning face attributes in the wild.
\newblock In {\em Proceedings of International Conference on Computer Vision
  (ICCV)}, December 2015.

\bibitem{Mescheder2018WhichTM}
Lars~M. Mescheder, Andreas Geiger, and Sebastian Nowozin.
\newblock Which training methods for gans do actually converge?
\newblock In {\em ICML}, 2018.

\bibitem{miyato2018spectral}
Takeru Miyato, Toshiki Kataoka, Masanori Koyama, and Yuichi Yoshida.
\newblock Spectral normalization for generative adversarial networks.
\newblock In {\em International Conference on Learning Representations}, 2018.

\bibitem{nizan2020breaking}
Ori Nizan and Ayellet Tal.
\newblock Breaking the cycle-colleagues are all you need.
\newblock In {\em Proceedings of the IEEE/CVF Conference on Computer Vision and
  Pattern Recognition}, pages 7860--7869, 2020.

\bibitem{oord2018representation}
Aaron van~den Oord, Yazhe Li, and Oriol Vinyals.
\newblock Representation learning with contrastive predictive coding.
\newblock {\em arXiv preprint arXiv:1807.03748}, 2018.

\bibitem{park2020cut}
Taesung Park, Alexei~A. Efros, Richard Zhang, and Jun-Yan Zhu.
\newblock Contrastive learning for unpaired image-to-image translation.
\newblock In {\em European Conference on Computer Vision}, 2020.

\bibitem{kmodal-stylegan}
Omry Sendik, Dani Lischinski, and Daniel Cohen-Or.
\newblock Unsupervised k-modal styled content generation.
\newblock {\em ACM Trans. Graph.}, 39(4), July 2020.

\bibitem{NIPS2016_6200}
Kihyuk Sohn.
\newblock Improved deep metric learning with multi-class n-pair loss objective.
\newblock In D.~D. Lee, M. Sugiyama, U.~V. Luxburg, I. Guyon, and R. Garnett,
  editors, {\em Advances in Neural Information Processing Systems 29}, pages
  1857--1865. Curran Associates, Inc., 2016.

\bibitem{wu2018unsupervised}
Zhirong Wu, Yuanjun Xiong, X~Yu Stella, and Dahua Lin.
\newblock Unsupervised feature learning via non-parametric instance
  discrimination.
\newblock In {\em Proceedings of the IEEE Conference on Computer Vision and
  Pattern Recognition}, 2018.

\bibitem{CycleGAN2017}
Jun-Yan Zhu, Taesung Park, Phillip Isola, and Alexei~A Efros.
\newblock Unpaired image-to-image translation using cycle-consistent
  adversarial networks.
\newblock In {\em Computer Vision (ICCV), 2017 IEEE International Conference
  on}, 2017.

\end{thebibliography}
}

\end{document}